\documentclass{article}
\pdfoutput=1 
\usepackage[a4paper, total={5.8in, 8.25in}]{geometry}
\usepackage{booktabs} 
\usepackage{subcaption}
\usepackage{tikz,xcolor,hyperref}

\definecolor{lime}{HTML}{A6CE39}
\DeclareRobustCommand{\orcidicon}{
	\begin{tikzpicture}
	\draw[lime, fill=lime] (0,0) 
	circle [radius=0.16] 
	node[white] {{\fontfamily{lmss}\selectfont \tiny \href{https://orcid.org/0000-0002-7972-1649}{ID}}};
	\draw[white, fill=white] (-0.0625,0.095) 
	circle [radius=0.007];
	\end{tikzpicture}
}
\foreach \x in {A, ..., Z}{\expandafter\xdef\csname orcid\x\endcsname{\noexpand
			{\noexpand\orcidicon}}
}

\hypersetup{%
    pdfborder = {0 0 0}
}
    \usepackage{tikz}
    \usetikzlibrary{shapes.geometric, arrows, positioning, automata}
    \tikzstyle{process} = [rectangle, minimum width=2cm, minimum height=1cm, text centered, draw=black, fill=white!30]
    \tikzstyle{sum} = \tikzstyle{sum} = [draw, circle, minimum size=.5cm]
    \tikzstyle{arrow} = [thick,->,>=stealth]

\usepackage{amsmath}

\usepackage{bm}
\usepackage{units}
\usepackage{float}
\usepackage{dblfloatfix}
\usepackage{hyperref}

\author{
  Fuda van Diggelen\\
  \small{Vrije Universiteit Amsterdam}\\
  \href{mailto:fuda.van.diggelen@vu.nl}{\texttt{\small{fuda.van.diggelen@vu.nl}}}\hspace{-0.25em}\orcidA{}
  \and
  E. Ferrante\\
  \small{Vrije Universiteit Amsterdam}\\
  \href{mailto:e.ferrante@vu.nl}{\texttt{\small{e.ferrante@vu.nl}}}
  \and
  A.E. Eiben\\
  \small{Vrije Universiteit Amsterdam}\\
  \href{mailto:a.e.eiben@vu.nl}{\texttt{\small{a.e.eiben@vu.nl}}}
}
\date{}

\begin{document}
\title{Comparing lifetime learning methods for morphologically evolving robots}
\maketitle
\begin{abstract}
Evolving morphologies and controllers of robots simultaneously leads to a problem: Even if the parents have well-matching bodies and brains, the stochastic recombination can break this match and cause a body-brain mismatch in their offspring. We argue that this can be mitigated by having newborn robots perform a learning process that optimizes their inherited brain quickly after birth. We compare three different algorithms for doing this. To this end, we consider three algorithmic properties, efficiency, efficacy, and the sensitivity to differences in the morphologies of the robots that run the learning process. 
\end{abstract}
\section{Introduction}
Evolutionary Robotics (ER) is concerned with using evolutionary methods to automate the design process of robots \cite{nolfi2000evolutionary,bongard2013evolutionary,vargas2014horizons, doncieux2015evolutionary}. In general, robots consist of two major components, a body (morphology, hardware) and a brain (controller, software) for which evolution can be employed to optimize both. As of today, however, a majority of ER studies considers the evolution of brains with a fixed body. \cite{winfield2015evolvable}. The specific morphologies encountered in the literature vary from simple wheeled robots, like e-pucks or Thymios, to robots with two, four or more legs, but the usual approach is to take one or more of such robots, task(s) and environment(s) and evolve controllers for high task performance. 

We deem this approach too limited as it can only answer the question \emph{What is the best controller for this robot?} for some task and environment. However, when evolving both the morphology and controller we can answer the question \emph{What is the best robot?} Despite such potential advantages, the simultaneous evolution of morphologies and controllers has not received much attention from the research community so far. 

One of the possible reasons is that jointly evolving morphologies and controllers can be very challenging \cite{lipson2016difficulty}. A key issue noted long ago \cite{eiben2013triangle} is that the stochastic nature of reproduction can lead to a body-brain mismatch problem: Even though parents have well-matching bodies and brains, recombination and mutation can shuffle the parental genotypes such that the resulting body and brain combination will not fit well. Consequently, causing sub-optimal behaviour in the offspring. The proposed solution is the addition of lifetime learning. As phrased in \cite{Eiben-Hart-2020} ``If it evolves it needs to learn''.

A generic system architecture to implement this solution is the Triangle of Life framework that integrates evolution and lifetime learning \cite{eiben2013triangle}. The essence is to have newborn robots perform a learning process that optimizes their inherited brain quickly after birth. An important additional feature is that newborn robots are considered infertile (i.e., not eligible for reproduction) until they successfully finish the learning period. This protection prevents the propagation of inferior older individuals over newer designs, which saves resources.

The main goal of this paper is to identify specific challenges to learning methods applied to newborn robots and investigate several possible algorithms for this. The two most important things to note here are the following. First, an evolutionary process will produce a large variety of morphologies. The shapes, sizes, and complexity of the evolved robots can be very different and unpredictable. Consequently, a suitable learning algorithm needs to work well on all possible morphologies, or at least be moderately sensitive for morphological differences. Second, the number of trials a learning algorithm is allowed to spend works as a multiplier of the total computational effort. For example, when the evolutionary algorithm has a budget of 10.000 fitness evaluations (e.g., 100 generations of populations of 100 robots) and each newborn robot can learn through 1000 trials\footnote{We maintain a broad view and see all generate-and-test search methods as potential learning algorithms if they can work in the `brain space' of the robots at hand.} then the total number of search steps is 10.000.000. This implies a strong preference for learning with a very low budget. 

The specific objective of this study is to compare three different algorithms that have been successfully applied as learners in robots:
\begin{enumerate}
    \item Bayesian Optimization \cite{lan2020learning},
    \item Evolutionary Strategy \cite{le2020sample},
    \item Differential Evolution \cite{tomczak2020differential},
\end{enumerate}  
\noindent With a tight budget of 300 trials. The comparison will take three measures into account: efficiency, efficacy, and the robustness to different morphologies.

\section{Related Work}
The body-brain mismatch problem can lead to early convergence with low variation of morphologies in the population \cite{lipson2016difficulty}. 
Several papers have suggested slightly different solutions to mitigate this effect on the population.
Cheney \textit{et al.} \cite{cheney2018scalable} implemented a form of novelty protection in which `younger' robot designs were protected from individuals that survived for more generations. Protecting a novel individual will increase its chance to adapt the controller properly for its body. Novelty protection corresponds with implementing a single lifetime learning iteration every time a morphology is protected. Similarly, De Carlo \textit{et al.} \cite{de2020influences} implemented protection in the form of speciation within their NEAT algorithm. The preservation of diversity in the population allowed new morphologies to survive, thus reducing the effects of body-brain mismatch. 

Nygaard, Samuelsen, and Glette \cite{nygaard2017overcoming} demonstrated improvements in their ER system by introducing two phases during evolution. The first phase consists of both controller and morphology evolution, while during the second phase only the controller evolves in a fixed body. The results showed that, without the second phase, morphology and controller evolution led to sub-optimal controllers which required additional fine-tuning. Lifetime learning is similar to the second phase but occurs more frequently after every recombination.

Last but not least, let us clarify the relation between lifetime learning as we discuss it here and existing ER research on controller evolution. Optimizing robot controllers by evolutionary algorithms on given, fixed robot morphologies (the majority of ER research) can be seen as a sub-area within the field of robot controller optimization. In that field, the focus is on the optimization of a specific robot with a possibly high computational budget and ER is distinguished by the choice for a specific type of optimizers, Evolutionary Algorithms. For our study, robot evolution is the enveloping context, where an evolutionary process optimizes robot morphologies and controllers simultaneously and we use lifetime learning as an addition within the evolutionary loop. As said above, we are very open and see many types of algorithms as potential learners, e.g., Reinforcement Learning, Simulated Annealing, Bayesian Optimization and Evolutionary Algorithms.



\section{System description}

\subsection{Robot morphologies}
Our robots consist of modules based on the RoboGen framework \cite{auerbach2014robogen}. We altered the original design to match real-world counterparts and only use the following modules: core component, brick components, and active hinges. The core component function as a container for a micro-controller with battery, and is a large brick with four attachment slots on its lateral faces. The brick component is a smaller cube with attachment points on its lateral sides. The active hinge component is a joint actuated by a servomotor with an attachment point on both ends. We can describe a robot's genotype using a tree structure with the core module as the root node.

\subsection{Robot controllers} 
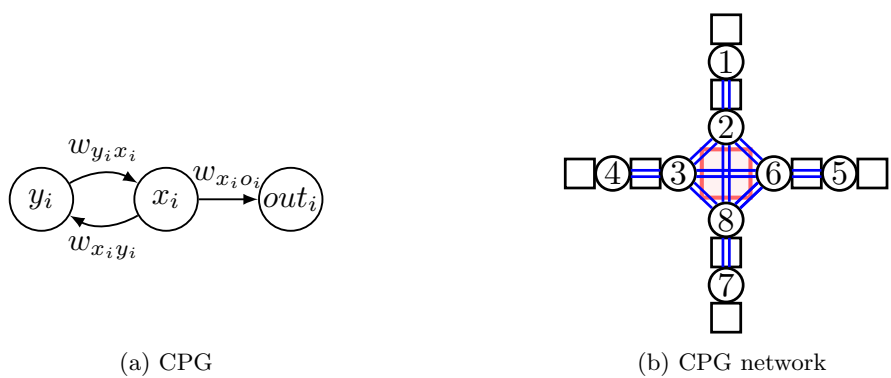
\begin{figure}[h]
\centering
\begin{minipage}[c]{\linewidth}
        \begin{minipage}[t]{0.5\linewidth}
            \centering
            \subfloat[CPG]{\centering
\resizebox{0.6\linewidth}{!}{
    \begin {tikzpicture}
	    [-latex,
		auto,
		node distance =1.375cm and 1.375cm,
		on grid,
		semithick,
		state/.style = {circle,draw,inner sep=0pt,minimum width = 0.7 cm}]
	\node[state] (X) {$x_i$};
	\node[state] (Y) [left = of X] {$y_i$};
	\node[state] (Output) [right = of X] {$out_i$};
	\path (X) edge [bend left = 30] node [] {$w_{x_iy_i}$} (Y);
	\path (Y) edge [bend left = 30] node [] {$w_{y_ix_i}$} (X);
	\path (X) edge  node [] {$w_{x_io_i}$} (Output);
	\node (ghost) [below of = X]       {};
	\end {tikzpicture}
}}
        \end{minipage}
        \hfill
        \begin{minipage}[t]{0.5\linewidth}
            \centering
            \subfloat[CPG network]{\centering
\resizebox{0.6\linewidth}{!}{%
\begin {tikzpicture}
	    [
		auto,
		node distance =0.3375cm and 0.3375cm,
		on grid,
		thick,
		state/.style = {circle,draw,inner sep=0pt, minimum width = 0.35 cm},
		box/.style={rectangle,draw,minimum width = 0.3 cm, minimum size=0.3 cm},
		core/.style={rectangle,draw=red!60,fill=red!5,very thick,minimum width = 0.5 cm, minimum size=0.5 cm}]

	\node[box] (B_l1) {};
	\node[state] (O4) [right = of B_l1] {$4$};
	\node[box] (B_l2) [right = of O4] {};
	\node[state] (O3) [right = of B_l2] {$3$};
	\node[core] (core) [right = 0.49cm of O3] {};
	\node[state] (O2) [above=0.48cm of core] {$2$};
	\node[box] (B_t2) [above=of O2] {};
	\node[state] (O1) [above=of B_t2] {$1$};
	\node[box] (B_t1) [above=of O1] {};
	\node[state] (O6) [right= 0.49cm of core] {$6$};
	\node[box] (B_r2) [right= of O6] {};
	\node[state] (O5) [right=of B_r2] {$5$};
	\node[box] (B_r1) [right=of O5] {};
	\node[state] (O8) [below=0.48cm of core] {$8$};
	\node[box] (B_b2) [below=of O8] {};
	\node[state] (O7) [below=of B_b2] {$7$};
	\node[box] (B_b1) [below=of O7] {};
	\path[draw=blue] (O4.10) to node [] {} (O3.170);
	\path[draw=blue] (O3.-170) to node [] {} (O4.-10);
	\path[draw=blue] (O6.10) to node [] {} (O5.170);
	\path[draw=blue] (O5.-170) to node [] {} (O6.-10);
	\path[draw=blue] (O2.100) to node [] {} (O1.-100);
	\path[draw=blue] (O1.-80) to node [] {} (O2.80);
	\path[draw=blue] (O7.100) to node [] {} (O8.-100);
	\path[draw=blue] (O8.-80) to node [] {} (O7.80);
	\path[draw=blue] (O3.50) to node [] {} (O2.-140);
	\path[draw=blue] (O2.-120) to node [] {} (O3.30);
	\path[draw=blue] (O3.10) to node [] {} (O6.170);
	\path[draw=blue] (O6.-170) to node [] {} (O3.-10);
	\path[draw=blue] (O3.10) to node [] {} (O6.170);
	\path[draw=blue] (O6.-170) to node [] {} (O3.-10);
	\path[draw=blue] (O6.150) to node [] {} (O2.-60);
	\path[draw=blue] (O2.-40) to node [] {} (O6.130);
	\path[draw=blue] (O2.-80) to node [] {} (O8.80);
	\path[draw=blue] (O8.100) to node [] {} (O2.-100);
	\path[draw=blue] (O6.-130) to node [] {} (O8.40);
	\path[draw=blue] (O8.60) to node [] {} (O6.-150);
	\path[draw=blue] (O6.-130) to node [] {} (O8.40);
	\path[draw=blue] (O8.60) to node [] {} (O6.-150);
	\path[draw=blue] (O3.-30) to node [] {} (O8.120);
	\path[draw=blue] (O8.140) to node [] {} (O3.-50);
	\end {tikzpicture}
}}
        \end{minipage}

           \vspace{-.5em} \caption{\textbf{a}) A single CPG. \textbf{b}) The CPG network for our Spider, containing 8 CPGs (numbers) with 10 connections (blue lines).}\label{fig:CPG_model}
\end{minipage}
\end{figure}

In this study, the controllers are optimized for gait-learning. We used a network of interconnected Central Pattern Generators (CPGs) with internal model control feedback, as it has shown to improve learning \cite{van2020effects}. In short, CPGs are pairs of artificial neurons ($x_i$,$y_i$) that reciprocally inhibit and excite each other to produce oscillatory behaviour \cite{ijspeert2008central}. Here, $i$ denotes the specific joint that is associated with a single CPG. The change of a neuron state is calculated by multiplying the current state of the opposite neuron with a weight ($w$), shown in \autoref{fig:CPG_model}a. The weights $w_{x_iy_i}$ and $w_{y_ix_i}$ represent a connection between each neuron pair. A output neuron $out_i$ is coupled to the $x$-neuron with an strength $w_{x_io_i}$. The output of $out_i$ will serve as the input command to the servomotors in a joint (\textit{i.e.} servo). In our case, we picked a tangent hyperbolic activation function as our output neuron to bound the motor input between [-1,1].

To enable more complex output patterns, CPG connections between neighbouring joints are allowed, see \autoref{fig:CPG_model}b. We define neighbouring joints at joint $i$ as the set ($\mathcal{N}_i$) of each joint $j$ positioned within a Manhattan distance of 2 modules. The resulting differential equations of the CPG network is shown in \autoref{eq:ODE_CPG}. Connections between neighboring joints $j$ are denoted as $w_{x_jx_i}$.
        
        \begin{align}\label{eq:ODE_CPG}
                \Dot{x}_i = w_{y_ix_i}y_i + \sum_{j \in \mathcal{N}_i} x_{j} w_{x_jx_i} &&\Dot{y}_i = w_{x_iy_i}x_i
        \end{align}
        
By changing the weights of the CPGs, we can improve our controller for the task of gait-learning. To simplify the search space, the following relations are imposed between the weights: $w_{x_iy_i} = -w_{y_ix_i}$; $ w_{x_jx_i} = -w_{x_ix_j}$; $w_{x_io_i} = 1$. At the start of each learning trial all neuron states will be set to a predefined value $(x,y)=\left(-\frac{1}{\sqrt{2}},\frac{1}{\sqrt{2}}\right)$. The sampling period of the controller is $\unit[0.125]{s}$.
        
        
\subsection{Learning methods}

As mentioned before, gait-learning is done by changing the weights of their respective CPG-network. We limit the number of evaluation to 300 for all learners as it has shown to be sufficient to reach convergence with BO \cite{lan2020time}.

\textit{Bayesian Optimization} is a state-of-the-art framework that had successful implementations in machine-learning, engineering, and science \cite{jasper2012practical}. In short, the BO algorithm contains two different functionalities. The first functionality is a function approximator that tries to model the fitness as a function of the search space parameters (in our case the weights of the CPG network). In theory, this approximator can be any type of function, but in BO Gaussian Processes ($\mathcal{GP}$) are used for their sample efficiency. The second functionality is an acquisition function that selects the next sample in the search space. The acquisition function chooses between samples that show high potential in fitness by the $\mathcal{GP}$ (exploitation), or that show a high degree of uncertainty by the $\mathcal{GP}$ (exploration). 

With the specific choice on the type of kernel for the $\mathcal{GP}$ and acquisition function, we can tailor the BO algorithm. We choose an upper confidence bound acquisition function and Mat\'ern 5/2 kernel. As initialization, we use Latin Hypercube Sampling that selects 50 evenly distributed pseudo-random samples in search-space. The hyper-parameters of the BO are shown in \autoref{tab:parameters} (from \cite{lan2020learning}).

\textit{Evolutionary Strategies} are evolutionary optimization algorithms in which a mutation operation is done by adding a random vector in search space. The distribution of the random vector (\textit{i.e.} mean, variance and correlation with a multivariate distribution) is often defined using Co-variance Matrix adaptation (CMA-ES). A basic cycle of the CMA-ES is as follows: (1) draw $\lambda$-samples from a normal distribution over search space; (2) evaluate the fitness over $\lambda$-samples; (3) perform selection over the top $\mu$-samples; (4) With the new population adapt the covariance matrix with $\sigma$-step (that decreases over the generations); (5) draw $\lambda$-samples according to the new distribution defined by the covariance matrix and repeat from (1).

For our ES learner, we implement a special CMA-ES algorithm with Novelty search and restart mechanisms with Increasing Population (NIPES), from \cite{le2020sample}. Here, novelty is based on the difference in CPG-weights between individuals within the current population and an archive of previous generations. Restart mechanisms re-initializes the evolutionary process with a doubled population size, every time a stopping criterion is met. These mechanisms address early convergence, the degeneration of the covariance matrix and the need for more global exploration. Our initial population size ($\lambda_0$) is set at 10 individuals, with $\sigma$ equal to 1. Stagnation threshold is defined as a standard deviation $\leq 0.05$ over the fitness values within a population and the between the last 5 generations. Other hyper-parameters of the ES algorithm are shown in \autoref{tab:parameters}.

\textit{Differential Evolution} is an EA that samples new candidates by perturbing the current population. This perturbation is calculated using linear operations over two genotype vectors that are obtained by randomly shuffling the current population. An advantage of this approach is that the optimization does not require any assumption on the distribution of the population (unlike ES), meaning that the DE approach is model agnostic. A potential limitation in DE is that it can suffer from small populations and loose diversity too quickly. This is because the randomness of the perturbation (when creating new samples) is directly linked to the size of the population. Unfortunately, we can not afford a very large population size to prevent this limitation, due to our tight budget of 300 evaluations. 

Reversible DE (RevDE) proposes a solution by applying a special set of linearly reversible transformation to produce a wide spread over search space when selecting new samples \cite{tomczak2020differential}. The algorithm works as follows: (1) initialize a population with $\lambda$-samples (2) evaluate the fitness over all $\lambda$-samples; (3) perform selection over the top samples to obtain genotype vector $\mu_1$; (4) randomly shuffle $\mu_1$ twice to obtain two additional vectors ($\mu_2,\mu_3$), and create new samples with the following linear relations: $\lambda_1=\mu_1+F(\mu_2-\mu_3)$, $\lambda_2=\mu_2+F(\mu_3-\lambda_1)$,  $\lambda_3=\mu_3+F(\lambda_1-\lambda_2)$. Apply uniform crossover with probability $CR$ between each $\lambda_n$ and $\mu_n$, and repeat from (2). The number of evaluations per generation$(\lambda$-samples) is 3 times that of $\mu_1$ which set to 10 (\autoref{tab:parameters}). Scaling factor $F=0.5$ has shown to result in stable exploration/exploitation behaviour in RevDE, with a $CR=0.9$ (from \cite{tomczak2020differential}). 

\begin{table}[ht!]
    \setlength{\textfloatsep}{0.7\baselineskip plus 0.2\baselineskip minus 0.5\baselineskip}    
    \small
    \caption{Hyperparameters of each learning algorithm}
    \label{tab:parameters}
    \centering
    \renewcommand{\arraystretch}{1}
    \begin{tabular}{l ll}
    \toprule
    \textbf{BO}         & Value & Description \\ \midrule
    Initial sampling    & LHS       & Sampling method\\
    Initial samples     & 50        & Number of samples\\ 
    Learning iterations & 250       & Number of evaluations\\ 
    Kernel type         & Mat\'ern 5/2 & Approximation kernel\\
    Kernel variance     & 1.0       \\ 
    Kernel length       & 0.2       \\ 
    UCB alpha           & 3.0       & Acquisition function weight\\  
    \bottomrule
    \toprule
    \textbf{NIPES}         & Value     & Description \\ \midrule
    $\lambda$           & 10        & initial population \\
    $\sigma$            & 1.0       & Adaption step size\\ 
    Stagnation threshold & $\leq 0.05$ & STD over fitness \\ 
    Stagnation window size & 5 & Number of generations\\
    Novelty: $\mu$      & 1         & Initial ratio\\
    Novelty: $\mu$-decrement & 0.05 &   \\
    Novelty: k          & 15        & Nearest neighbours \\
    Novelty: threshold  & 0.9       & To add to archive\\
    Novelty: probability & 0.4      & To add to archive\\
    \bottomrule
    \toprule
    \textbf{RevDE}         & Value     & Description \\ \midrule
    $\lambda$           & 30        & Population size \\
    $\mu$               & 10        & Top-samples size  \\
    $F$                 & 0.5       & Scaling factor\\ 
    $CR$                & 0.9       & Crossover probability \\ 
    \bottomrule
    \end{tabular}
\end{table}


\section{Experiments}

A link to the git repository can be found here\footnote{\url{https://github.com/fudavd/revolve/tree/learning}}. The straightforward way to compare the three learning algorithms is to plug them into our evolutionary robot system and run a good number of repetitions with different random seeds for each combination. However, this would require an extremely long time to conduct. (With 300 learning trials per newborn robot it would take 300 times longer than the pure evolutionary run.) Therefore, we introduce an alternative methodology, loosely based on the standard way of comparing machine learning algorithms. 

The idea is to create a test suite of several robot morphologies and test the learning algorithms on the robots in the test suite only. This is not unlike evaluating machine learning algorithms on different types of specific data sets. A given robot and a given data set play a similar role as they represent a particular learning problem to be solved. Thus, the overall logic is similar, the learning methods are compared on a suite of specific problems (problem instances), $N$ robots or $N$ data sets. To make a robust assessment of the performance per data set, we implement multiple repetitions ($N=30$) with random initialization over these robots.

\begin{figure*}[hb!]
\input{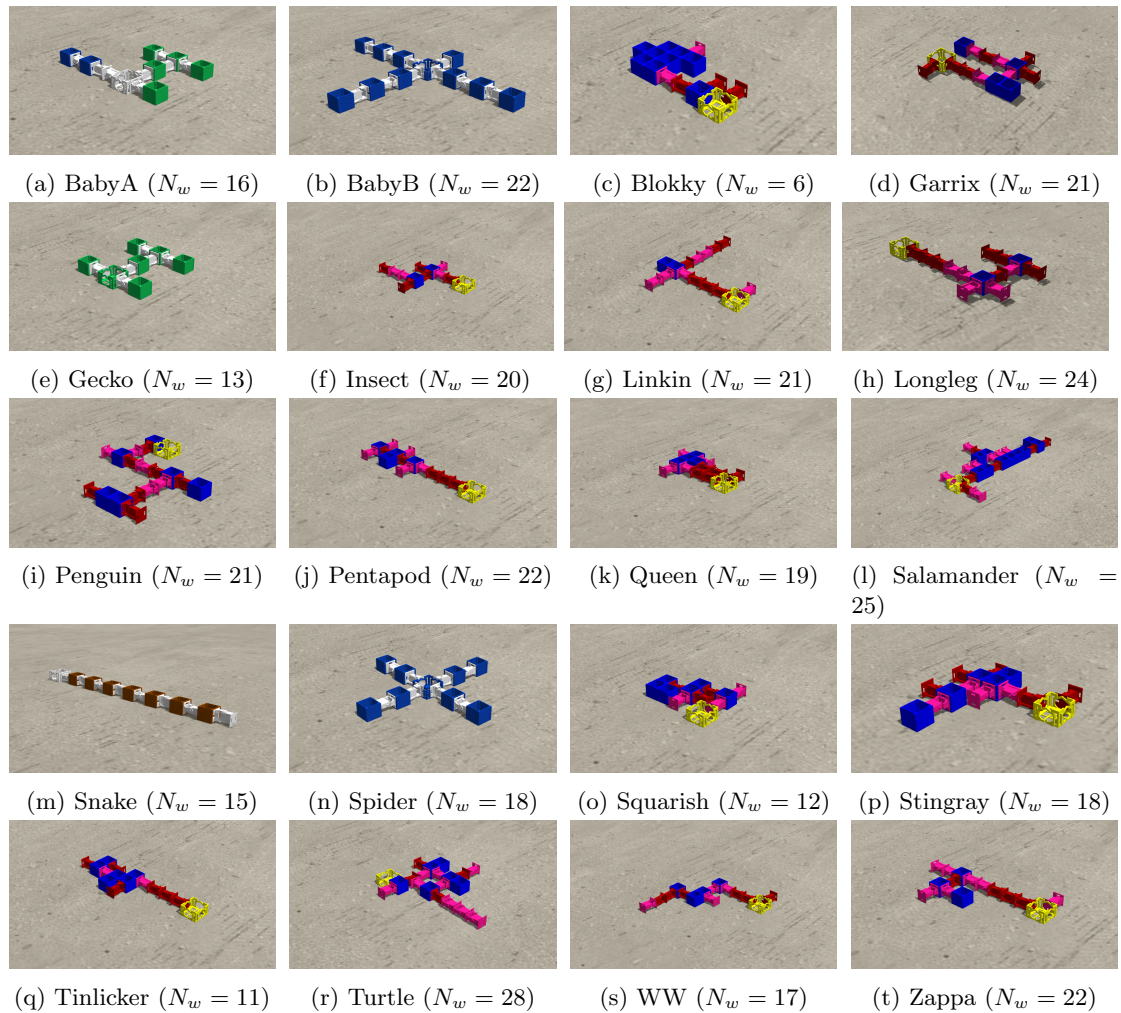}
\caption{\label{fig:test_suite}Test suite of all 20 robots. The base set consisted of three hand-designed robots (Spider, Gecko and Snake), with 2 children robots (BabyA and BabyB). The remaining set was selected from a population of evolved robots, using novelty search.}
\end{figure*}

\subsection{Test suite}
For a sizeable and diverse test suite, we selected 20 morphologically different robots based on seven morphological traits. Specifically, \textit{size}, \textit{number of joints}, \textit{proportion}, \textit{branching}, \textit{number of limbs}, \textit{coverage}, and \textit{symmetry} from \cite{miras2018search}. We created the test suite in two stages. First, we started from five robots: three hand-designed Gecko, Spider, and Snake, and two `children' (from Gecko and Spider), BabyA and BabyB. The other 15 robots were selected by an algorithm from a population of 100 individuals. These individuals were evolved with a fitness function that combined speed and novelty \cite{miras2018effects}. The algorithm that selected the additional 15 robots maximized the pairwise differences in the 7-dimensional space, spanned by the morphological traits mentioned above. The resulting set of 20 robots is shown in \autoref{fig:test_suite}.

\subsection{Learner performance measures}
To compare the different learning algorithms, we consider three performance indicators: \textit{efficiency}, \textit{efficacy}, and the \textit{robustness to different morphologies}. The assessments are based on inspecting the learning curves for each morphology and each learning algorithm. Additionally, to evaluate the overall performance we will aggregate the results over the morphologies ($N=20$) and perform statistical analysis (one-way ANOVA) to see if there is a significant difference between the learning methods (for $p\leq0.05$). An additional post-hoc analysis between the learners is done if we find a significant difference. Violations of the assumption on normality and/or equal variance result in taking a different appropriate statistical measure. The specific definitions of the performance indicators are as follows. 

\textit{Efficacy}: We measure efficacy by the quality achieved at the end of the learning process. Since we consider gait learning here, the quality of a learner is defined by the speed of the robot's best controller. As this measure can be sensitive to `luck', we get more useful statistics by taking the average over 30 different runs. Thus here, the efficacy of a learning method is defined by the mean best fitness (MBF) averaged over the 30 independent repetitions.

\textit{Efficiency}: Efficiency indicates how quickly the learner finds its best solution. To this end, we use the Average number of Evaluations to Solution (AES) as defined in \cite{eib2003introduction}. More than indicating efficiency, AES can provide insight into the maximum number of evaluations required for our learner. If we overestimate the maximum number of evaluations required, then the AES values tend to be (much) lower compared to the evaluation limit set during the experiment.

\textit{Robustness}: The robustness of a learner is defined by the differences in performance over various robot morphologies. We can measure this in two different ways, by the variance of the learners' efficacy (ROB-MBF) or the variance of the learners' efficiency (ROB-AES) calculated over the test suite. 
Significant differences in variance between the groups can be found by conducting an $F$-test of equality of variances (with $\mathrm{df=(2, 57)}$). 

Adding to the above measures, it is interesting to consider the standard deviation of the MDF and AES values over the 30 repetitions on individual robot morphologies. We call this \textit{consistency} as it provides information on the sensitivity of an algorithm to `luck'. A learner with a small standard deviation is consistent in its performance, while larger values show that it can get lucky or unlucky easily. Similarly to robustness, we distinguish consistency in quality (CON-MBF) and consistency in efficiency (CON-AES).  

On top of the statistical comparison, we are also interested in the shape of the fitness landscapes. Implementing novelty search in the selection process of our 20 robots resulted in 20 data points spread out over morphological space \cite{miras2018effects}. We approximate the (learner specific) fitness landscape by interpolating the fitness values between each of these data points. To this end, we use a Gaussian mask to calculate a weighted average over each point in morphological space (see \autoref{eq:Gauss}). The interpolation works as follows. For each data point $k$ (20 robots) a Gaussian ($\mathcal{P}_k(x,y)$) is place at the corresponding position $(x_k, y_k)$ in morphological space, where dimensions $x$ and $y$ refer any combination of two morphological traits. Subsequently, a weighted average over all fitnesses is taken at each point $x$ and $y$ to create a full contour plot as shown in \autoref{eq:Weight}. Here, $F_k$ denotes the MBF value of robot $k$.

\begin{equation}\label{eq:Gauss}
    \begin{aligned}
        \mathcal{P}_k(x,y) &= \frac{1}{2 \pi \sigma_p^2}exp\Bigg\{-\frac{(x-x_k)^2}{2\sigma_{p}^2} -\frac{(y-y_k)^2}{2\sigma_p^2}\Bigg\}\\
    \end{aligned}
\end{equation}
\begin{equation}\label{eq:Weight}
    \begin{aligned}
        \mathcal{F}(x,y) &= \frac{\sum_k \mathcal{P}_k(x,y)F_k}{\sum_k \mathcal{P}_k(x,y)}\hphantom{xxxxxxxxXxxlxlxlxl}\\
    \end{aligned}
\end{equation}

This method creates a smooth landscape with a very local effect of each fitness. The locality is solely dependent on the shape of the Gaussian mask, parameterized with $\sigma=0.1$. It should be noted that this contour plot, at most, provides an insight into the shape of the fitness landscape and does not produce a valuable fitness estimation. We, therefore, present the height of our fitness landscape with minimum-maximum bounds. Nevertheless, the contour plots help us to identify different attractors in morphology space for each learner.

\subsection{Test procedure} 
The three learning algorithms (BO, ES, and DE) are compared with the 20 robots in our test suite. To make a reliable assessment, we will conduct $30$ different runs on each robot per learner (so $30\times3\times20=1800$ runs in total). Each run consists of $300$ learning trials (i.e. evaluation) and has no other termination condition. At the start of each learning trial, we place a robot in neutral position (all joint angles $\unit[0]{rad}$) centred on a flat horizontal plane with a gravitational pull of $\unit[9.81]{m/s^2}$. We use Gazebo for the simulation with the ODE physics engine and RK4 integration ($\mathrm{dt}=\unit[0.05]{s}$). For the task of gait-learning, we define the robot's fitness as its average speed in \unit[60]{s}, i.e. absolute distance in centimetres per second (\unit[]{cm/s}). 

\section{Results}
\begin{figure*}[ht!]
\input{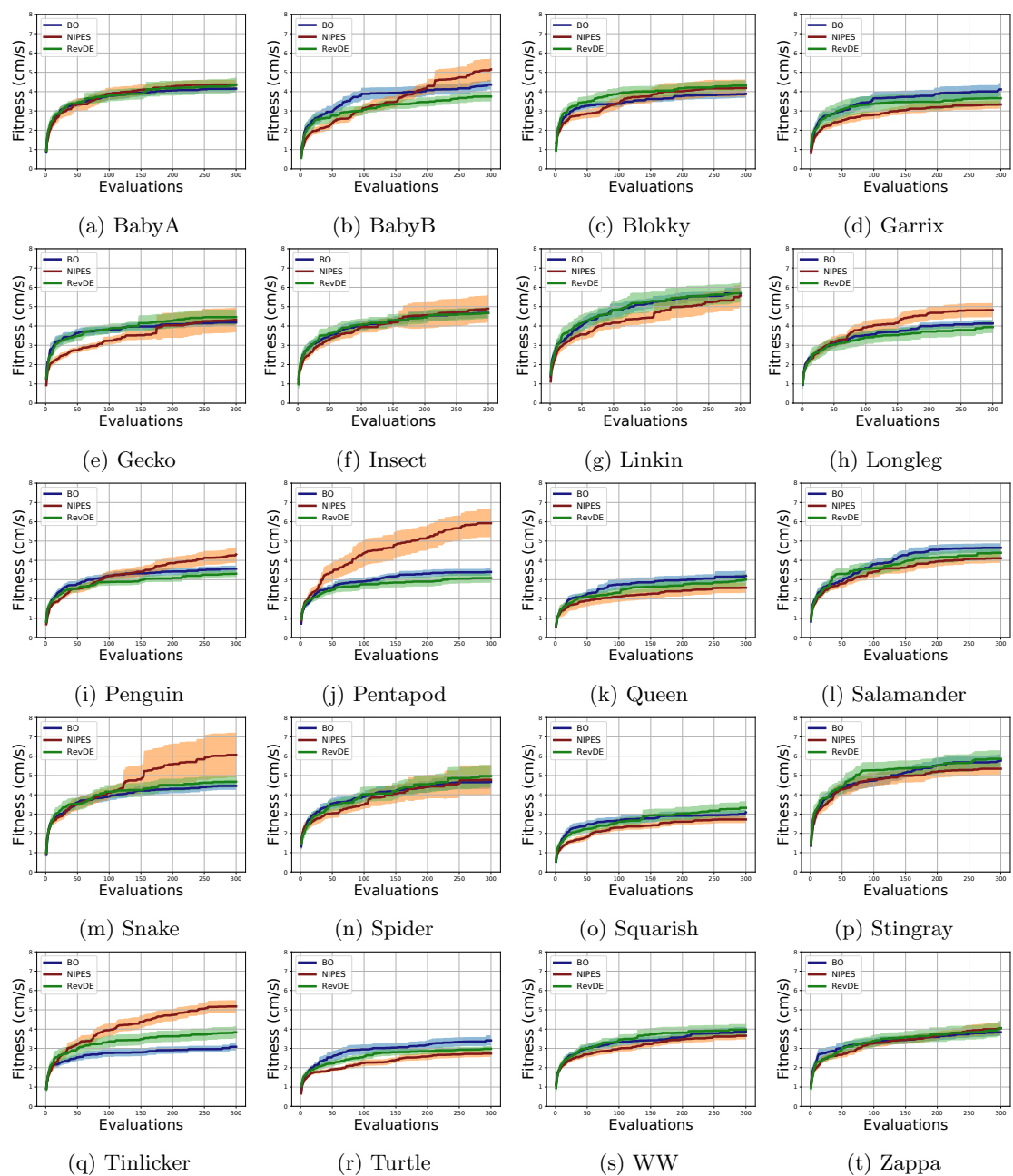}
\caption{Speed over time (the number of evaluations) during gait-learning, averaged over 30 independent repetitions. The blue lines show the Bayesian Optimization, red lines the evolution strategy (NIPES) , and the green lines the differential evolution (RevDE). The bands indicate the 95\% confidence intervals ($\pm1.96\times\mathrm{SE}$).}
\label{fig:res_curve}
\end{figure*}

All three algorithms could successfully learn good controllers on all morphologies. \autoref{fig:res_curve} exhibits the fitness curves for BO in blue, NIPES in red, and RevDE in green, showing the mean best fitness values over 30 runs with their respective 95\% confidence intervals ($\pm1.96\times SE$, Standard Error). A significant difference between the curves can be identified when there is no overlap between the curves. In 8 cases there is no significant difference between BO, ES and DE, for the BabyA, Gecko, Insect, Linkin, Spider, Stingray, WW, and Zappa. NIPES significantly outperforms the other learners on six robots (BabyB, Longleg, Penguin, Pentapod, Snake, and Tinlicker), BO only outperforms on Turtle, while RevDE never beats the other two learners with a significant margin. 

Based on our data we can identify the `champions', i.e., the fastest robots in our test suite. Considering the top speed the Snake is the fastest (\unit[17.76]{cm/s}), followed by the Insect (\unit[14.71]{cm/s}), and the Spider (\unit[14.33]{cm/s}). Considering the MBF results, the best robots are the Snake (\unit[6.07]{cm/s}), Pentapod (\unit[5.93]{cm/s}), and Spider (\unit[5.86]{cm/s}). 

\begin{figure*}[h]
\input{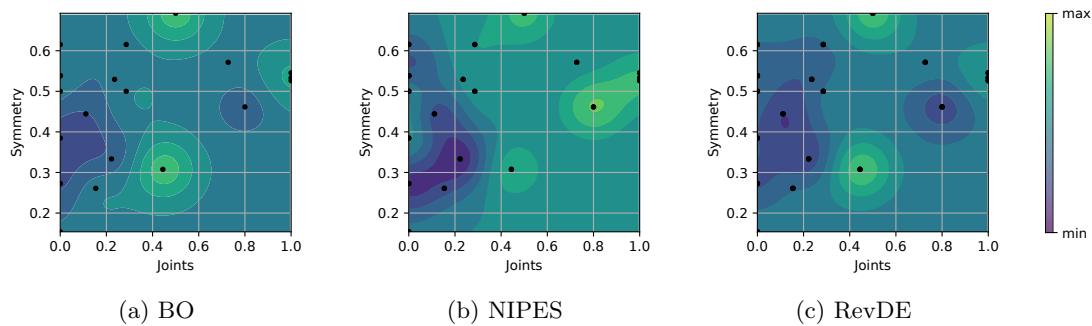}
\caption{\label{fig:res_contour}Fitness landscape based on symmetry and number of joints for each learner BO, NIPES and RevDE.}
\end{figure*}

Following up the obtained efficacy results, we can now produce our fitness landscapes (\autoref{fig:res_contour}). We only present a single instance of the possible morphological measure combinations, i.e. showing the fitness landscape as a function of \textit{joints} and \textit{symmetry}. These morphological measures have been used before to produce fitness landscape \cite{miras2018effects, de2020influences}, and show interesting differences between the learners. As mentioned before, the height in these plots only indicates possible attractors. Here, we can see that overall BO and RevDE seem to have similar attractors in morphological betweenspace, meaning that these learners perform well on similar types of morphologies. In comparison, NIPES seems to have a much larger basin of attraction and a deeper region of lower fitnesses. Furthermore, differences between the lowest and the highest point are more pronounced in NIPES compared to BO and RevDE, which correspond to the results from the fitness curves. Lastly, the shape of the fitness landscape in NIPES hints to a possible bias for morphologies with many joints. This could be an unwanted property for the implementation of lifetime learning.

We present a numerical summary of the results on the learner performance measures in \autoref{tab:res_DATA}. Assumptions on normality were met for all condition and measure combinations. Regarding \textit{efficacy}, we can see that the overall differences in the mean best fitness are not significant ($p>0.05$). On the measure of \textit{efficiency}, we did find significantly different results between the learners. It should be noted that the assumption of equal variances was violated here (visible in the significant difference of ROB-AES). Therefore, a more conservative Kruskal-Wallis test was performed to measure differences between the mean AES. With post-hoc analysis (using \textit{Mann–Whitney U}-test with Bonferroni correction), NIPES turns out to be slower than the other methods, since its AES values are significantly higher compared to both BO ($p<0.001$) and RevDE ($p<0.01$). In contrast, the AES of the BO and RevDE do not differ significantly from one another. 

\begin{center}
\begin{table}[h]
\small
    \caption{Efficacy (MBF), efficiency (AES), robustness (ROB-MBF and ROB-AES), and consistency (CON-MBF and CON-AES) per learning algorithm over the test suite of $\bm{N}$=20 different morphologies. The last column show the $\bm{p}$-values of each test, $\bm{F}$-test for VAR, and ANOVA for the other measures. The * indicates the use of Kruskal–Wallis test.
    \label{tab:res_DATA}}
    \centering
    \renewcommand{\arraystretch}{1.1}
    \begin{tabular}{l cccr}
    \toprule
    Measure                     & BO        & NIPES & RevDE &$p$-value  \\ \midrule
    MBF     & 4.11$\pm0.73$  & 4.40$\pm1.0$\hphantom{0} & 4.12$\pm0.77$ & 0.48 \\
    AES                         & 149$\pm10$\hphantom{0} & 171$\pm21$\hphantom{0} & 151$\pm13$\hphantom{0} & *$\leq$0.001 \\
    ROB-MBF         & 0.54 & 1.0 & 0.63 & 0.48 \\
    ROB-AES                         & 96 &456 &173 & $\leq$0.001 \\
    CON-MBF    & 0.66$\pm0.16$   & 1.24$\pm0.70$ & 0.86$\pm0.28$ & $\leq$0.001\\
    CON-AES    & 83$\pm9$\hphantom{0} & 75$\pm9$\hphantom{0} & 85$\pm6$\hphantom{0} & $\leq$0.001\\
    \bottomrule
    \end{tabular}
\end{table}
\end{center}

Considering \textit{robustness} we can distinguish the (in)sensitivity of the efficiency and the efficacy of the learners with respect to the different morphologies. The ROB-MBF show that the variance between the efficacy of learners does not differ significantly ($p>0.05$). In contrast, the robustness of the efficiency (ROB-AES) did differ significantly between the different learners ($p<0.001$). A closer inspection of this result reveals that ROB-AES in NIPES is significantly higher compared to BO ($p<0.001$) and RevDE ($p<0.05$), while BO and RevDE did not differ significantly. This indicates that the efficiency of NIPES varies a lot between different morphologies.

In contrast to the results on robustness to different morphologies, differences between the consistency in solution quality are significant. This difference in results for ROB-MBF and CON-MBF show that both measures do represent distinct properties of the learners. In particular, for NIPES the CON-MBF results show statistically significant differences ($p<0.001$) indicating that this learner is more subject to `luck' than BO ($p<0.001$) and DE ($p<0.001$). On a given robot the maximum speed can greatly vary for NIPES. This is not surprising taking into account that it is a greedy optimizer that can get trapped in local optima. The low budget of 300 trials is apparently not enough to compensate for this effect by restarts. Furthermore, CON-AES shows that NIPES has significantly less variation compared to BO ($p<0.01$) and RevDE ($p<0.001$). This shows that between the variation between runs within the same morphology is lower. 

Together these results provide important insights into the overall quality of the learners. In the end, some differences in terms of sample efficiency can be seen, as the BO and RevDE deliver the same solution quality (walking speed for the robot, MBF) with fewer evaluations than NIPES (based on AES). In addition, two indirect insights follow the results on robustness and consistency: 1) A full ER experiment (evolving both morphology and controller) with NIPES as a lifetime learner, will perform significantly more inconsistent (i.e. luck-based) than BO or RevDE. This inconsistency will be more pronounced as we omit the repeated runs per morphology with lifetime learning. 2) We successfully negated the effect of `luckiness' in NIPES by repeating 30 runs per morphology, meaning that we converged sufficiently to the true MBF. If we did not, we would also suspect to see statistically significant differences in variance as well (ROB-MBF). Furthermore, NIPES is less robust than the other two algorithms, because its efficiency varies more over different robot morphologies.

\section{Discussion}
Recall that the algorithms used in our experiments are meant to be used in combination with morphological robot evolution as learning methods in newborn robots. One of the contributions of this paper is the methodology to compare such learning algorithms without running the full-blown experiments, meaning evolution with learning. The validity of this methodology is motivated by three ideas. Firstly, we can show differences between learners with several specific robots as test cases. The straightforward approach to compose an adequate test suite would be to select a ``representative'' set of robots. However, it is not clear what ``representative'' should mean in our context. This motivates the second idea: we chose to create a set of robots that is well-spread in the space of possible morphologies and we define that space through several quantifiable morphological traits. In this paper we used seven traits, leading to a 7-dimensional space with a natural distance measure. Lastly, each individual in the test set should have enough potential to learn. Comparison is only informative when learners can differentiate from one another. We achieved this by selecting from a previously evolved population.

It is important to note that changing the design space of the robots changes the applicable morphological traits. For instance, if the design allows wheels, then the number of wheels of a robot is an appropriate trait that defines a new dimension of the morphological space. In general, it is not known what traits are relevant for the working of the learning algorithms. For example, the size of a robot may be important and the proportion of its length and width irrelevant. Thus, our recommendation is to use all possible traits until more research on the long term delivers insights to select more cleverly. 

 Our view on applicable learning algorithms is very permissive: we consider any generate-and-test style search algorithm applicable, if only it can search in the space of possible controller configurations. Obviously, this depends on the given controller architecture and the learnable components therein, but in general, the number of options is large, including, for instance, Bayesian Optimisation (that we considered here), Reinforcement Learning and Simulated Annealing (that we did not), and all kinds of evolutionary algorithms (of which we studied two). Let us briefly note that if the learning algorithm is `accidentally' evolutionary, then great care needs to be taken to avoid confusion with the evolutionary method employed to evolve the robots. By the lack of space, we cannot discuss this in detail here. Let us only note that it is useful to distinguish the outer evolutionary loop, where new individuals are new robots with a body and a brain and the inner evolutionary loop (the learning method), where new individuals are new brains for the given robot body.

A given test suite of robots with a number of learning algorithms can be compared by the usual metrics regarding efficiency, efficacy, and robustness. Rooted in our application, there are two very important measures: \emph{efficiency} because infant robots should optimize their controllers quickly after birth; and \emph{robustness}, because evolution (the outer loop) can produce very different robots and the learner should work well on all of them. The current study is the first of its kind into this, which means that we should be careful with generalisations. However, our data suggest that finding robust learners is not too hard and that 300 learning trials --probably even 200-- can be enough to achieve decent performance. Additionally, a recent study indicates that the learning potential of evolved robots is growing throughout evolution \cite{Miras2020a}. If this turns out to be a general effect, then the maximum number of learning trials could even be lowered. The obvious caveat is that we only considered one task in a simple environment. More complex tasks and environments may require more steps and may induce more pronounced differences among learning algorithms.

\section{Conclusions and future work}
In this paper, we investigated learning methods to be applied in newborn robots within evolutionary robot systems, where morphologies and controllers evolve simultaneously. We conducted an empirical comparison of three methods, Bayesian Optimization, an Evolution Strategy NIPES and a Differential Evolution RevDE on a test suite of 20 robots for the task of gait learning. The results show that BO and RevDE can deliver the same solution quality (walking speed for the robot) with fewer evaluations than NIPES. Furthermore, NIPES is less robust than the other two algorithms, because its performance varies more over different robot morphologies. 

Ongoing work is concerned with extending the scope of this study by investigating other tasks and other, more complex environments including underwater applications. A significant extension of the current setup will consider multiple tasks for the evolving robots besides locomotion.

\bibliographystyle{abbrv}
\bibliography{bibliography} 

\end{document}